\documentclass[letterpaper, 10 pt, conference]{ieeeconf}  

\IEEEoverridecommandlockouts                              

\usepackage{graphics} 
\usepackage{epsfig} 
\usepackage{mathptmx} 
\usepackage{times} 
\usepackage{amsmath} 
\usepackage{amssymb}  
\usepackage{float}
\usepackage{afterpage}
\usepackage{changepage}
\usepackage{multicol}
\usepackage{makecell} 
\usepackage[font=small,labelfont=bf]{caption}

\usepackage{subcaption}
\usepackage{adjustbox}
\usepackage{colortbl}
\usepackage{lipsum} 
\usepackage{microtype}
\usepackage{booktabs}
\usepackage{graphicx}
\usepackage{multirow}
\usepackage{pifont}
\usepackage{url}
\usepackage{xparse}
\usepackage{xspace}
\usepackage{nicefrac}
\usepackage{enumerate}
\usepackage{pifont}%
\usepackage{setspace}
\usepackage{wrapfig}
\usepackage{xfrac}
\usepackage{tikz}
\usepackage{stfloats}
\usepackage[most]{tcolorbox}
\usepackage{hyperref}
\usepackage{etoolbox}

\newcommand{\xmark}{\ding{55}}
\newcommand{\algabbr}{EgoMI\xspace}
\newcommand{\memabbr}{SPARKS\xspace}

\title{\LARGE \bf
\algabbr: Learning Active Vision and Whole-Body Manipulation from Egocentric Human Demonstrations

}

\author{
Justin Yu$^{1,2*\dag}$\protect\thanks{$^{*}$ Equal contribution}\protect\thanks{$^{\dag}$ Work done during internship at xdof.ai.}, Yide Shentu$^{1,2*}$, Di Wu$^{2}$, Pieter Abbeel$^{1}$, Ken Goldberg$^{1}$, \\ Philipp Wu$^{2}$ \\
$^{1}$UC Berkeley, $^{2}$xdof.ai 
}

\begin{document}

\makeatletter
\let\@oldmaketitle\@maketitle%
\renewcommand{\@maketitle}{\@oldmaketitle%
    \centering
    \vspace*{-0.8ex}
    \includegraphics[width=1.0\linewidth]{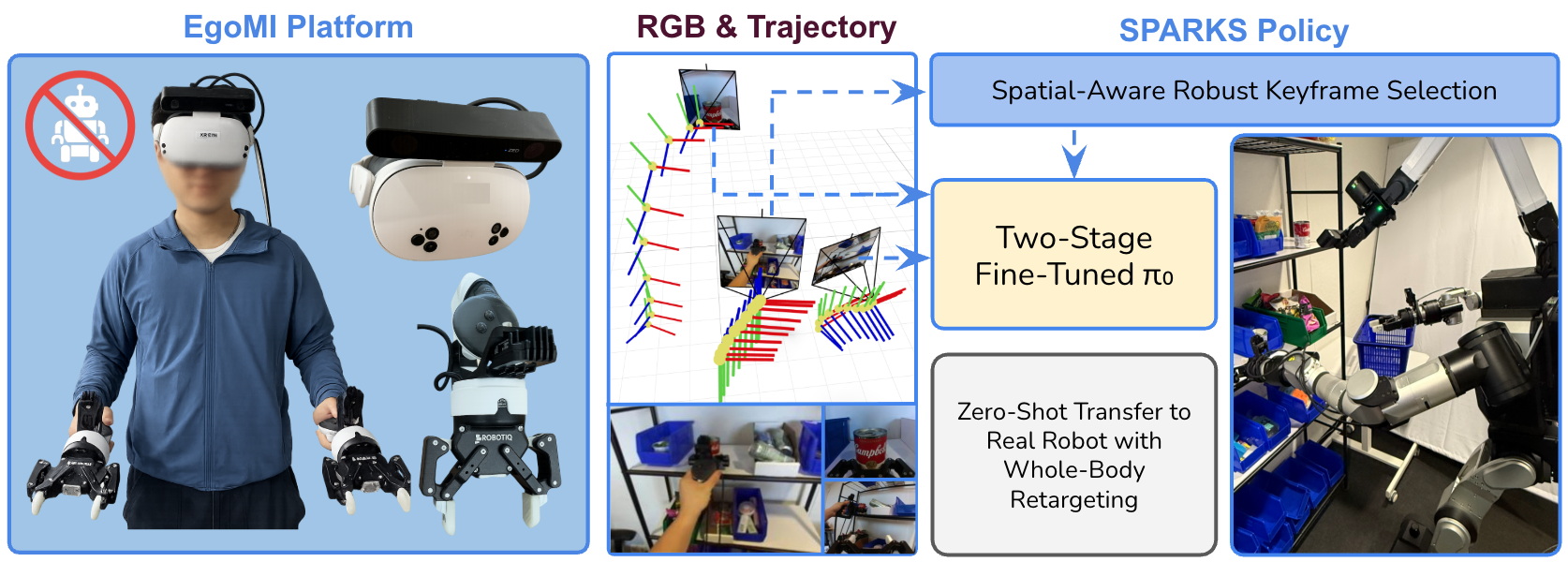}
    \vspace*{-0.5ex}
    \captionof{figure}{\textbf{Overview of the \algabbr framework.} 
    \algabbr captures egocentric human demonstrations with synchronized head and hand tracking. To handle rapid viewpoint changes from head motion, demonstrations are processed via \emph{Spatial-Aware Robust Keyframe Selection} (\memabbr). A two-stage fine-tuning procedure then adapts a pre-trained absolute joint-space foundation model ($\pi_0$) into \emph{Relative-Operation Space}. The learned policy remarkably transfers zero-shot to real robots through whole-body retargeting, without requiring any visual augmentation, explicit visual alignment, or on-embodiment data collection.
    }
    \label{fig:teaser}
    \vspace*{-2.5ex}
}
\makeatother

\maketitle
\thispagestyle{empty}
\pagestyle{empty}

\begin{abstract}
Imitation learning from human demonstrations offers a promising approach for robot skill acquisition, but egocentric human data introduces fundamental challenges due to the embodiment gap. During manipulation, humans actively coordinate head and hand movements, continuously reposition their viewpoint and use pre-action visual search strategies to locate task-relevant objects. These behaviors create dynamic, task-driven head motions that static robot sensing systems cannot replicate, leading to a significant distribution shift that degrades policy performance. We present \algabbr (Egocentric Manipulation Interface), a framework that captures synchronized end-effector and active head trajectories during manipulation tasks, resulting in data that can be retargeted to compatible semi-humanoid robot embodiments. To handle rapid and wide-spanning head viewpoint changes, we introduce a memory-augmented policy that selectively incorporates context from historical observations. We evaluate our approach on a bimanual robot equipped with an actuated camera head and find that policies with explicit head-motion modeling consistently outperform baseline methods. Results suggest that coordinated hand–eye learning with \algabbr effectively bridges the human-robot embodiment gap for robust imitation learning on semi-humanoid embodiments. Project page: \url{https://egocentric-manipulation-interface.github.io}
\end{abstract}



\section{INTRODUCTION}

Learning from large-scale human demonstrations represents a powerful pathway toward scalable robot skill acquisition. Recent advances in imitation learning have shown promising results in training robots directly from human collected data. However, a significant barrier persists in the form of the \emph{embodiment gap}, the fundamental mismatch between human demonstrators and robotic platforms.
During manipulation tasks, humans actively integrate head and eye movements with hand actions, dynamically adjusting their viewpoint to maintain visual contact with task-relevant objects and resolve occlusions. This active perception strategy is fundamental to how we understand and interact with our environment.

In contrast, many contemporary robotic systems rely on static-external camera configurations that cannot replicate this coordinated visual behavior. This creates a severe distribution shift when learning from egocentric data, as the task-driven viewpoint changes inherent in human demonstrations cannot be reproduced by fixed sensing systems. Various methods attempt to minimize the embodiment gap through strategies such as restricting to wrist mounted cameras \cite{9126187, chi2024universal, shafiullah2023bringing, young2020visualimitationeasy} or projecting top camera views into a coordinate-invariant representation \cite{chen2024arcapcollectinghighqualityhuman, wang2024dexcap}. However, for more complex tasks that require searching or looking with the head, the learned policy is unable to reproduce the demonstrated behavior. Furthermore, the inability of standard policies to maintain a persistent spatial memory of past observations exacerbates this gap, leading to context loss during rapid head movements.

To address this challenge, we introduce the Egocentric Manipulation Interface (\algabbr), as illustrated in \autoref{fig:teaser}, a framework that captures the full degrees of freedom of human perceptual movements with minimal embodiment gap. \algabbr simultaneously records both end-effector and head movements during human demonstrations. Proprioception and camera data streams for both the wrist and head result in more complete conditioning information for the downstream policy. We can then retarget a policy's prediction of the whole-body motion data onto a robot. We use a modified Rainbow-RBY1 (wheeled, semi humanoid robot), equipped with a 6-DoF arm that functions as a neck to enable faithful retargeting of human head and end-effector motions to the robot.

A primary challenge of utilizing an actuated head is the potential for context loss due to large viewpoint shifts. To address this, we present Spatial-Aware Robust Keyframe Selection (\memabbr), a lightweight algorithm that selects a compact set of past head keyframe images to mitigate the loss of context caused by rapid head viewpoint shifts, while avoiding out-of-distribution failures. \memabbr is a simple yet effective mechanism that emphasizes keyframes rich in spatial information, and scores past frames using viewpoint novelty, temporal recency, and motion smoothness as a training-free proxy for visual informativeness. By embedding spatial memory into robot policies, \memabbr enables more stable long-horizon reasoning and resilience to viewpoint shifts. 

Real world robot experiments show that incorporating head observations and movement from the egocentric demonstration is crucial during both the data collection stage and policy deployment stage. Policies trained without the head trajectory data or memory consistently fail due to a lack of critical context. Remarkably, our framework achieves zero-robot-data transfer \emph{without relying on any} augmentations, in-painting, or viewpoint re-rendering, demonstrating that capturing egocentric head and hand motions and observations is sufficient to bridge the embodiment gap. In summary, we:
\begin{itemize}
    \item Demonstrate the importance of an actuated head for imitation learning in everyday robotic manipulation tasks.  
    \item Introduce a simple yet effective approach for training robot policies with spatial memory, addressing the challenges posed by rapid perspective changes from the egocentric head camera.  
    \item Develop a data collection device that records key egocentric data.
    \item Evaluate the approach in real-world experiments, demonstrating capabilities enabled by head retargeting and memory-aware policies.
    \item Release code, hardware designs, and experiments to facilitate reproducibility and further research.
\end{itemize}


\section{Related Work}

\subsection{Data collection devices for imitation learning}
Progress in imitation learning is closely intertwined with advancements in the devices used to capture demonstrations, as the choice of interface heavily influences both the quality of collected data and scalability of the process. Devices used for teleoperation vary widely, ranging from joysticks and 3D spacemice~\cite{liu2022robot,zhu2022viola}, VR controllers~\cite{rakita2017motion,shridhar2023perceiver,arunachalam2023holo}, and camera based sensors~\cite{9126187, handa2020dexpilot,sivakumar2022robotic,qin2023anyteleop}. While effective for many tasks, these devices typically require control abstractions to be simplified into end-effector space due to mismatches between human morphology and robot kinematics.  
Leader–follower systems such as ALOHA~\cite{fu2024mobile}, GELLO~\cite{wu2023gello} and AirExo-2~\cite{fang2025airexo} mitigate this issue by matching the teleoperation device more directly to the robot morphology. UMI~\cite{chi2024universal} represents a middle ground, using only the robot gripper as the input device.
Large-scale datasets such as EgoDex~\cite{egodex-learning-dexterous-manipulation} exemplify this approach. However, these methods still exhibit a large embodiment gap, since human motions, perspectives, and physical morphology differ substantially from a robot’s own embodiment.


\subsection{Imitation learning approaches from human data}
Recent advances in large-scale imitation learning have resulted in robot foundation model approaches that integrate pre-trained vision and language representations with behavior cloning to scale imitation learning across numerous tasks ~\cite{Lynch2022-na, Jang2022BCZZT, brohan2022rt1, brohan2023rt2, lynch2021language}.
Follow-on systems such as RoboFlamingo~\cite{li2023vision}, Octo~\cite{octo_2023},  OpenVLA~\cite{kim24openvla} $\pi_0$~\cite{black2024pi0visionlanguageactionflowmodel}~\cite{driess2025knowledgeinsulatingvisionlanguageactionmodels}, $\pi_{0.5}$\cite{intelligence2025pi05visionlanguageactionmodelopenworld}, and Seed GR-3~\cite{cheang2025gr3technicalreport} confirm that combining demonstration data with foundation models yields strong zero-shot generalization and versatile open-world behavior.

While the above largely use teleoperated data, others focus on data from human egocentric data. One notable example is EgoMimic~\cite{kareer2024egomimicscalingimitationlearning}, which captures egocentric video and detailed 3D hand tracking using augmented reality (AR) glasses and trains policies on both human and robot data to improve generalization. DexCap~\cite{wang2024dexcap} introduces scalable finger-level demonstrations collected via motion-capture gloves alongside a specialized imitation algorithm to precisely capture fine manipulation details. Building upon their works, ARCap~\cite{chen2024arcapcollectinghighqualityhuman} harnesses AR feedback to guide novice operators in recording robot-executable trajectories, improving demonstration quality and usability. These efforts collectively push forward scalable and accessible high-fidelity data collection for imitation learning. Our approach additionally considers head motion and memory.

\subsection{Policy learning with Active Vision}

Policy learning with active vision seeks to endow agents with the ability to intelligently control their viewpoints dynamically, coordinating sensory actions (e.g., camera movements or gaze shifts) with motor actions to reveal task-relevant information and mitigate occlusions. Early approaches relied on fixed~\cite{Zhao2023LearningFB,fu2024mobile,shaw2024bimanual} or eye-in-hand cameras~\cite{10333330}, while more recent work combines reinforcement and imitation learning to jointly optimize perception and action.

ViA~\cite{xiong2025visionactionlearningactive} uses a 6-DoF head-mounted camera to imitate human head motion, reducing occlusions and selecting informative viewpoints, but leverages a robot embodiment specific teleoperation data collection system. EyeRobot~\cite{kerr2025eyerobotlearninglook} equips a robotic system with a 2DoF “eyeball” camera and learns gaze behavior through a combined BC-RL (behavior cloning and reinforcement learning) objective, while manipulation is learned through behavior cloning. Look, Focus, Act~\cite{chuang2025lookfocusactefficient} incorporates human gaze data into a foveated vision encoder, showing how explicit fixation signals improve policy robustness. In contrast, EgoMI directly captures human head motion for natural active vision and combines it with SPARKS, a lightweight memory mechanism to learn policies in a robot agnostic manner while still maintaining a minimal embodiment gap.

\section{\algabbr System}

In this section, we detail the \algabbr system, complete with the hardware device, data processing, and robot. The goal of the \algabbr design is to enable easy data collection for humans while minimizing the embodiment gap. Policies trained on \algabbr demonstrations transfer to semi-humanoid robotic embodiments \emph{without} requiring extensive data domain adaptation or the inclusion of teleoperated data on the embodiment of the deployed robot.



\begin{table}[t]
  \small
  \setlength{\tabcolsep}{3pt} 
  \centering
  \caption{Comparison of teleoperation systems and their features. 
    For on-embodiment data collection, teaching device accuracy is less critical since the robot records data from sensor feedback rather than depending on the fidelity of the teaching device itself. 
    Our proposed method, \algabbr, is the only system that simultaneously captures head and hand trajectories, supports true gripper actions, and enables whole-body retargeting, bridging the embodiment gap between human demonstrations and robotic execution. }
  \label{tab:teleop_comparison}
  \begin{tabular}{lcccc}
    \toprule
    System & \makecell{Error (mm) \\ (avg$_{\pm\text{std}}$)} & \makecell{Robot-free \\ Collection} & \makecell{Head Pose\\Tracking} & \makecell{True Gripper \\ Action} \\
    \midrule
    Gello   & N/A    & \xmark & \xmark  & \checkmark \\
    ALOHA   & N/A    & \xmark & \xmark  & \checkmark \\
    UMI     & $8.855_{\pm 3.228}$ & \checkmark & \xmark   & \xmark \\
    AirExo-2  & $1.737_{\pm 1.713}$ & \checkmark & \xmark      & \xmark \\
    VIA     & N/A & \xmark & \checkmark     & \checkmark \\
    \algabbr & $2.126_{\pm 1.216}$ & \checkmark & \checkmark     & \checkmark \\
    \bottomrule
  \end{tabular}
\label{data_collection_system}
\vspace*{-2ex}
\end{table}

\subsection{Data Collection Hardware}
\label{sec:hardware}

The \algabbr collection system integrates commercially available hardware with custom components to capture synchronized head, hand, and visual data in a format directly compatible with downstream robot execution, shown in Figure~\ref{fig:teaser}. The core of the system is a Meta Quest 3S VR headset, which provides 6-DoF tracking of the operator’s head and hand controllers. A camera (ZED 2i) is rigidly mounted above the headset to record first-person video that is aligned with head movements.
Each VR hand controller is augmented with two custom hardware features to further align with the robot embodiment. First, a mounting point for wrist cameras provides a view closely aligned with the robot’s wrist-mounted cameras for fine-grained manipulation tasks. 
Second, a mechanical flange interface with a standard mounting pattern allows the system to interface with off the shelf gripper systems (in our case, a Robotiq 2F-85). 

During data collection, the triggers on each VR hand controller are mapped to real-time drive-by-wire control of the robot gripper actuation. For a comparison of \algabbr with respect to other data collection systems, refer to \autoref{tab:teleop_comparison}. Notably, \algabbr enables robot free data collection, synchronized streams of head pose, hand trajectories, gripper action, proprioception, and egocentric and wrist videos, all while physically matching the geometry and visual appearance of the target robot platform.

\subsection{Operator Gaze and Active Vision Data}
While the \algabbr device captures operator head motion, one practical challenge is the absence of explicit eye-gaze tracking in our current hardware stack. Humans naturally fixate their gaze on task-relevant objects before acting~\cite{land1999roles}. To approximate this behavior, we overlay a fixed visual reticle at the center of the passthrough view and instruct operators to align it with manipulation targets and placement locations. This  addition imposes little cognitive load, formalizes natural gaze behavior, and enables head orientation to serve as a reliable proxy for the center of visual attention, following the human look-then-reach behavior observed in motor studies.

Centering provides significant benefits for downstream policies by driving task-relevant visual features into the middle of the observation space and creating an object-centric representation that couples perception and action. In contrast, fixed external-cameras or unactuated head-cameras scatter task-relevant visual features across the image plane, forcing models to rely on weaker positional encodings. Qualitatively, we observe that policies trained without the reticle often fail completely, likely due to free gaze variability and weak head–gaze correlation, which caused trajectories and observations to drift out of the demonstration distribution at deployment.



\subsection{Data Reformatting and Cleaning}
\label{sec:reformatting}
We develop a high-throughput conversion pipeline that discovers and validates trajectory episodes, filtering out low-quality or corrupted data via video timing checks, and trajectory smoothness thresholds (SE(3) translation/rotation deltas). A key step is applying transforms that re-orient the raw data to minimize the proprioceptive gap between the capture system and the target robot. Because demonstrations are collected in a VR system with its own arbitrary world frame, we align all poses to the robot’s canonical coordinate system based on the first time step data sample per-episode. This is done by applying a homogeneous transform that aligns the first timestep head position in the horizontal plane with the world-frame origin and re-orienting the forward-facing direction determined by the combined orientation of the gripper end-effectors. Specifically, let the raw VR-frame poses at time $t$ be
\[
T^{L}_{V}(t),\; T^{R}_{V}(t),\; T^{H}_{V}(t) \in SE(3) 
\]
where $R \in SO(3)$ and $p \in \mathbb{R}^3$.
Here, $L$ and $R$ refer to the left and right end-effectors, and $H$ to the head.

\subsubsection*{1) Forward direction estimation}
We first extract the forward-facing axis (the $z$-axes in our case) of the two end-effector frames at the first timestep:
\[
z_L(t) = R^{L}_{V}(t)\,e_z, \quad z_R(t) = R^{R}_{V}(t)\,e_z, 
\quad e_z = \begin{bmatrix}0 & 0 & 1\end{bmatrix}^\top.
\]
We project each vector onto the $xy$-plane and normalize:
\[
\bar{z}_L = \frac{\Pi_{xy} z_L(0)}{\|\Pi_{xy} z_L(0)\|}, 
\qquad
\bar{z}_R = \frac{\Pi_{xy} z_R(0)}{\|\Pi_{xy} z_R(0)\|},
\]
where the projection operator is
\[
\Pi_{xy} \begin{bmatrix} x \\ y \\ z \end{bmatrix} = 
\begin{bmatrix} x \\ y \\ 0 \end{bmatrix}.
\]

We then compute the yaw angles of each vector and take the circular mean to avoid discontinuities:
\[
\theta_L = \mathrm{atan2}(\bar{z}_{L,y}, \bar{z}_{L,x}), 
\quad
\theta_R = \mathrm{atan2}(\bar{z}_{R,y}, \bar{z}_{R,x}),
\]
\[
\theta = \mathrm{atan2}(\sin\theta_L + \sin\theta_R,\; \cos\theta_L + \cos\theta_R).
\]

We define this re-orientation component as a yaw rotation about the $z$-axis $R_z(\theta)$.

\subsubsection*{2) Base origin from head position}
We place the base-frame origin at the $x$-$y$ position of the head from the first timestep:
$t_B = \begin{bmatrix} p_{H,x} &  p_{H,y} &  0 \end{bmatrix}^\top.$

\subsubsection*{3) VR to base transform}
We define the base-to-VR and its inverse VR-to-base transform as:
\[
T^{B}_{V} =
\begin{bmatrix}
R_z(\theta) & t_B \\[4pt]
0 & 1
\end{bmatrix},
\
T^{V}_{B} = (T^{B}_{V})^{-1} =
\begin{bmatrix}
R_z(\theta)^\top & -R_z(\theta)^\top t_B \\[4pt]
0 & 1
\end{bmatrix}.
\]


\subsubsection*{4) Applying calibration and offsets}
Let $T^{L}_{\mathrm{flange}}$ and $T^{R}_{\mathrm{flange}}$ be the VR controller-to-robot flange calibration transforms,
and $T_{\mathrm{flange}\to\mathrm{TCP}}$ be the fixed tool-center-point (TCP) offset for the end-effectors.
The calibrated, base-frame end-effector poses are then
\[
T^{L/R}_{B}(t) = T^{V}_{B} \, T^{L/R}_{V}(t) \, T^{L/R}_{\mathrm{flange}} \, T_{\mathrm{flange}\to\mathrm{TCP}}
\]

For the head, let $T_{\mathrm{head}\to\mathrm{cam}}$ be the fixed transform from the VR headset to the camera:
\[
T^{H}_{B}(t) = T^{V}_{B} \, T^{H}_{V}(t) \, T_{\mathrm{head}\to\mathrm{cam}} 
\]

This process re-expresses all trajectories in a consistent robot-centric frame, minimizing the proprioceptive gap between VR-collected demonstrations and the target robot's embodiment.
\subsection{Robot Setup: Wheeled-Humanoid with Active Head}
\label{sec:platform}
To successfully transfer whole-body human movement to a robotic setup, it is necessary to have hardware that can simultaneously capture and reproduce the agility of both head and hand motion. Since humans naturally engage their waist, shoulders, and other joints during demonstrations, a robot system with more than a fixed torso and arms is necessary to realize such movements.

\begin{figure}[t]
  \centering
  \begin{minipage}[c]{0.60\linewidth}
    \includegraphics[width=\linewidth]{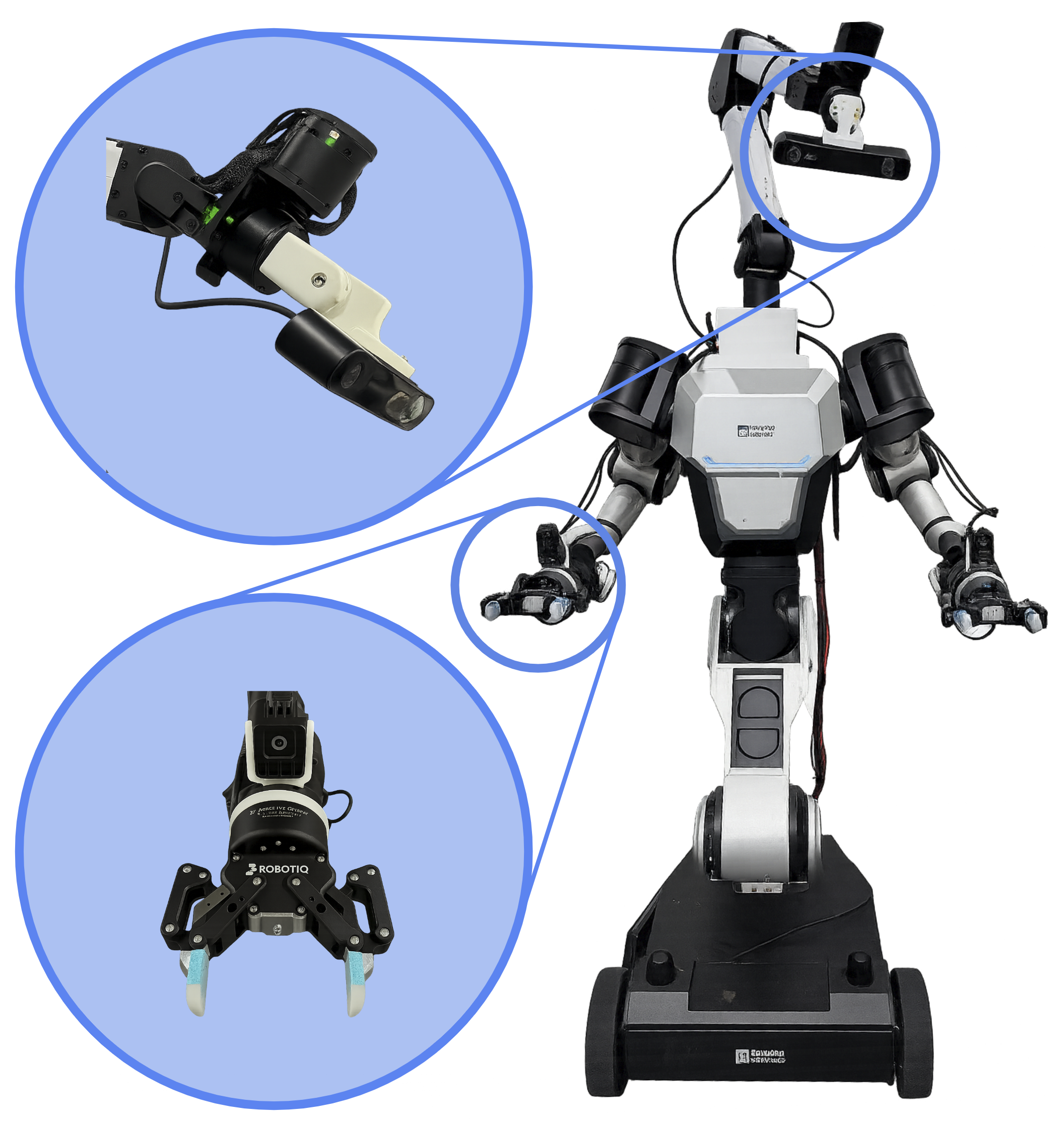}
  \end{minipage}%
  \hfill
  \begin{minipage}[c]{0.35\linewidth}
    \caption{\textbf{\algabbr policy deployment setup.} 
    We use a modified Rainbow RBY1 robot with a 6-DoF YAM \cite{i2rt_yam_arm} + ZED2i \cite{stereolabs_zed2i} camera mounted on top as the fully actuated head. The gripper configuration is identical to the human demonstration setup, minimizing the embodiment gap.}
    \label{fig:rby_sys}
  \end{minipage}
\vspace*{-3.0ex}
\end{figure}

Moreover, a fully actuated neck is required, as humans move their heads freely in 6D space during search and manipulation phases. For our experiments, we use a modified Rainbow RBY1 robot equipped with a 6-DoF torso and 2$\times$ 7-DoF arms. On top of its torso, we mount an I2RT YAM \cite{i2rt_yam_arm} robot with a ZED2i camera \cite{stereolabs_zed2i} as the active vision head, enabling us to simultaneously track and replicate the full 6DoF head and hand movements collected from \algabbr demonstrations. See Fig.~\ref{fig:rby_sys} for the detailed system setup.

\section{Method}

This section details the model training and inference approach to enable learning policies on \algabbr data. Furthermore, we detail Spatial-Aware Robust Keyframe Selection (\memabbr), as a simple but effective method to train policies with spatial memory.

\subsection{Policy Interface: 29D Action and State Representation}
\paragraph{Dataset action / state format (absolute, world-frame)}
In addition to synchronized image observations, each timestep of our dataset encodes a 29D \emph{action} vector and \emph{state} (proprioception) vector.
\[
\underbrace{\big[\;\overbrace{r^{L}_{6}}^{\text{rot6}},\; \overbrace{p^{L}_{3}}^{\text{pos}},\; \overbrace{g^{L}_{1}}^{\text{grip}},\;
\overbrace{r^{R}_{6}}^{\text{rot6}},\; \overbrace{p^{R}_{3}}^{\text{pos}},\; \overbrace{g^{R}_{1}}^{\text{grip}},\;
\overbrace{r^{H}_{6}}^{\text{rot6}},\; \overbrace{p^{H}_{3}}^{\text{pos}}\;\big]}_{\text{29D dataset vector}}
\]
where $r_{\cdot}^{(\cdot)} \in \mathbb{R}^{6}$ is a 6D rotation vector (first two columns or rows of the $\mathbb{R}^{3\times3}$ rotation matrix concatenated),
$p_{\cdot}^{(\cdot)} \in \mathbb{R}^{3}$ is position, and $g^{(\cdot)} \in \mathbb{R}$ is the continuous gripper signal.
For actions, $g^{L}, g^{R}$ come from operator intent (VR triggers); for state, they come from measured hardware closure/aperture.



\paragraph{Model input space (relative, inter-gripper)}
For model training, we keep the proprioception right hand in the world frame and compose the left hand and head as poses \emph{relative to the right}:
\[
{}^{R}\!T_{L} = \big({}^{W}\!T_{R}\big)^{-1}\, {}^{W}\!T_{L}, 
\qquad
{}^{R}\!T_{H} = \big({}^{W}\!T_{R}\big)^{-1}\, {}^{W}\!T_{H},
\]
We then parameterize $({}^{R}\!T_{L}, {}^{R}\!T_{H})$ by $(r^{L\!:\!R}_{6}, p^{L\!:\!R}_{3})$ and $(r^{H\!:\!R}_{6}, p^{H\!:\!R}_{3})$ using the same 6D rotation + 3D position ordering convention:
\[
\underbrace{\big[\; r^{L\!:\!R}_{6},\, p^{L\!:\!R}_{3},\, g^{L}_{1},\;
r^{R}_{6},\, p^{R}_{3},\, g^{R}_{1},\;
r^{H\!:\!R}_{6},\, p^{H\!:\!R}_{3}\;\big]}_{\text{29D proprio model input}}
\]
For action chunk samples, we transform from absolute to \emph{relative} parameterization over all SE(3) components excluding gripper actions, as introduced in~\cite{chi2024universal}.

\paragraph{Deployment space (absolute, world-frame)}
The inverse transforms reproject policy outputs back to absolute world-frame commands. Predicted 6D rotations are expanded into valid $3{\times}3$ rotation matrices via Gram-Schmidt orthonormalization.

\paragraph{$SO(3)$ Representation}
We choose this 6D vector representation for $SO(3)$ over other discontinuous representations such as Euler angles or double cover representations like quaternions, critical for the stability of neural network gradient-based optimization \cite{zhou2019continuity}.

\subsection{Spatial-Aware Robust Keyframe Selection (SPARKS)}
\label{sec:SPARKS}

Natural egocentric head motion produces rapid, task-driven viewpoint changes as the operator scans, fixates, and repositions their head during manipulation. Critical task information is often first revealed under viewpoints that differ dramatically from the current frame, such as when a human leans or turns their head to disambiguate occlusions or search for an object. 
In such settings, policies trained to only take a single timestep—or fixed windows—suffer from severe \emph{context loss}, as temporally distant but visually important observations are dropped from the model’s conditioning. 
\memabbr leverages the head trajectory to select a compact set of past frames for memory, avoiding costly learned or recurrent modules. In other words, \memabbr is a direct benefit of our active head sensing design, and would be neither necessary nor effective if the robot had only a static viewpoint.

At time $t$, given a causal head pose history in a short lookback window $\{\,{}^{W}\!T_H(\tau)\,|\,\tau \leq t\}$, \memabbr assigns past frame $\tau$ a score combining three factors:
\small\[
J(\tau)=
\underbrace{\phi\!\big(\angle(\hat{z}_H(\tau),\,\hat{z}_H(t))\big)}_{\text{viewpoint novelty}}
\;+\;
\underbrace{\psi(t-\tau)}_{\text{recency}}
\;+\;
\underbrace{\rho\!\big(\angle(\hat{z}_H(\tau-1),\,\hat{z}_H(\tau))\big)}_{\text{motion smoothness}},
\]
where $\hat{z}_H$ is the forward axis of the head camera transform.
The three terms encourage novel viewpoints, recency, and low angular velocity (avoiding blurred frames, along with the insight that frames are likely more informative when the operator's gaze is fixated).  
Only frames exceeding diversity thresholds (angular displacement $>\alpha \cdot \mathrm{FOV}$ or translation displacement $>\delta$) are added to a FIFO buffer.

\memabbr is run offline to precompute keyframe indices to preserve IID minibatch sampling during training, and run online at deployment in $O(L)$ per step, where $L$ is the lookback length.

\begin{figure*}[t]
  \centering
  \includegraphics[width=1.0\linewidth]{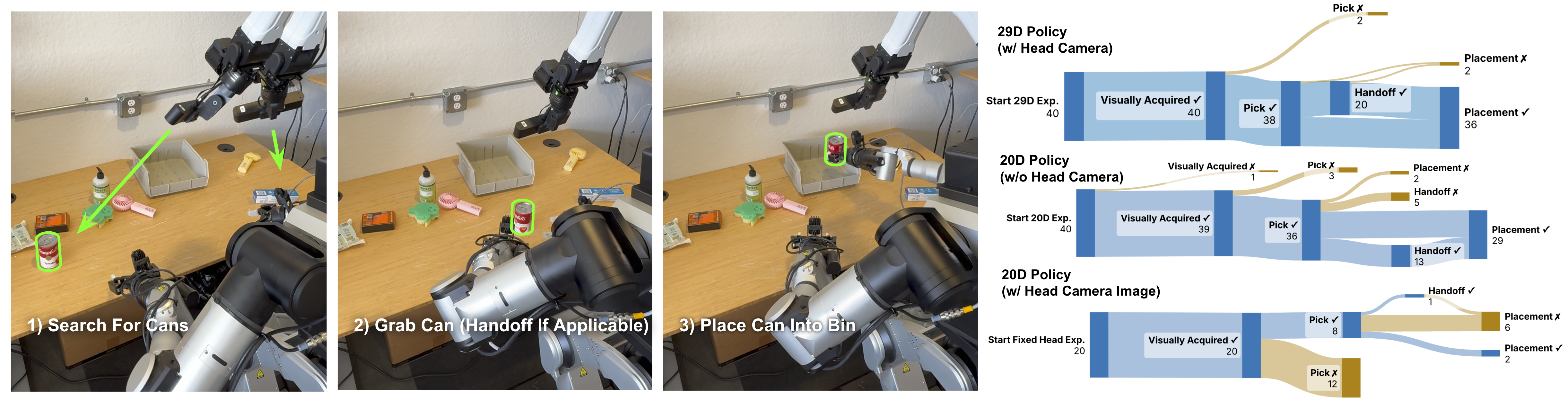}
  \caption{
  \textbf{Tabletop Task Rollout Sequence}: \textit{(Left).} The images show a real 29D policy evaluation rollout where the robot (1) scans for target cans across a cluttered workspace, (2) grasps the correct item with potential handoff between grippers, and (3) places it into the designated bin. \textit{(Right).} The Sankey diagrams illustrate failure modes between policies with full 29D action space and active head-camera versus reduced 20D wrist camera-only and 20D + head-camera images baselines.}
  \label{fig:envs4}
\end{figure*}
\begin{figure}[t]
  \centering
  \includegraphics[width=1.0\linewidth]{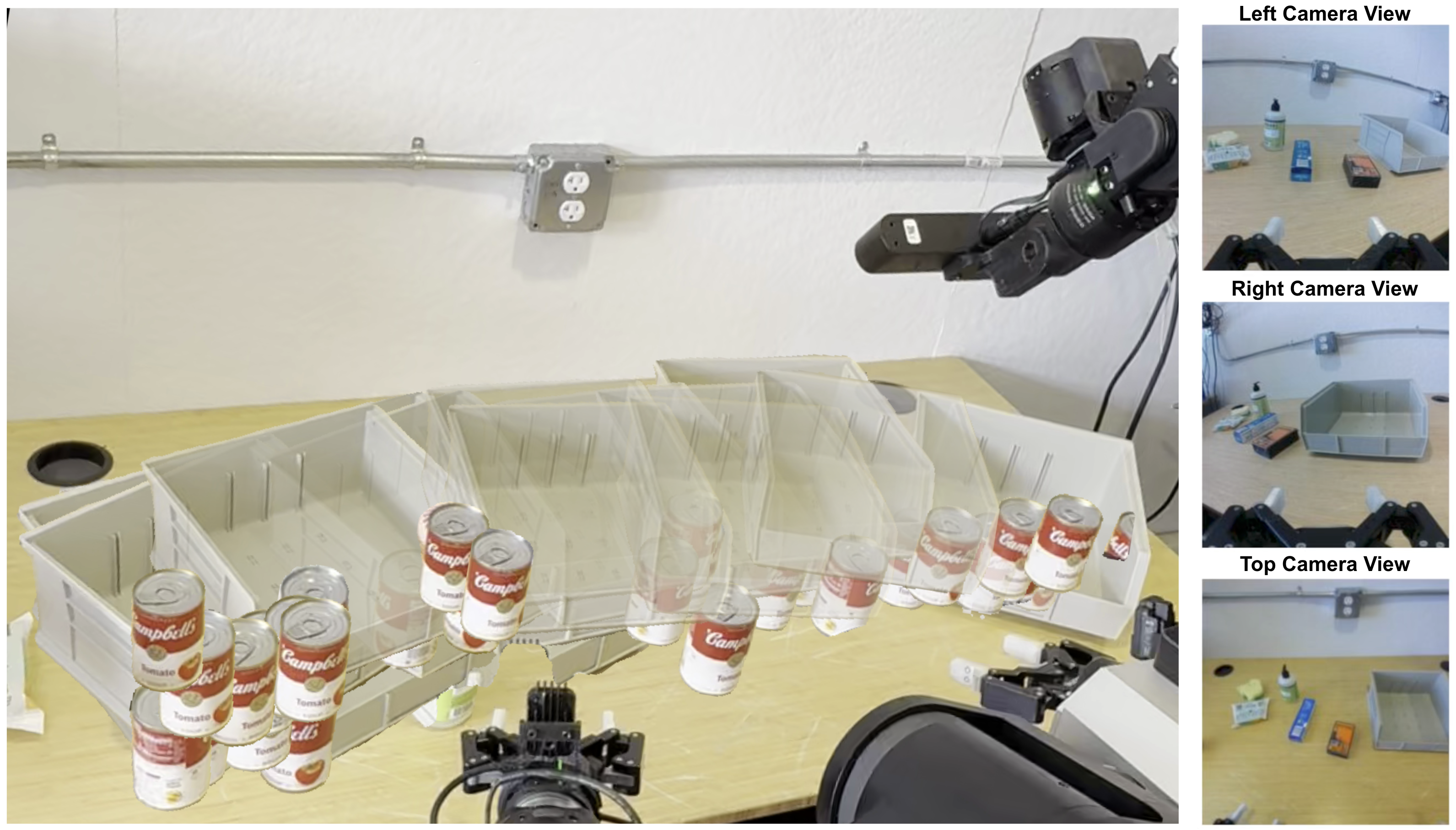}
  \caption{
  \textbf{Randomization distribution and example initial configuration of the tabletop environment} highlighting the wide distribution of object positions and clutter scenarios used during evaluation rollouts. \textit{(Right).} Initial configurations for target object may be outside of the immediate field of view of the initialized robot during experimentation. Target object and placement location may also reside on opposite ends of the workspace requiring a bi-manual handoff maneuver.
  }
  \label{fig:envs3}
\end{figure}

\subsection{Policy Training}
Starting with a capable pre-trained foundation model is crucial for achieving strong final performance. We initialize our approach from pre-trained $\pi_0$ model weights \cite{black2024pi0visionlanguageactionflowmodel}. However, $\pi_0$ was originally trained only on \textit{absolute robot joint positions}, which introduces a mismatch with our target representation. To bridge this gap and map $\pi_0$’s output space into the \textit{relative Cartesian space}, we adopt a \textbf{two-stage finetuning process}: 
\begin{enumerate}
    \item \textbf{General Multi-Task Finetuning to 29D action space.} We first finetune $\pi_0$ end-to-end on a diverse multi-task dataset from our in-house data bank, adapting $\pi_0$ from absolute joint outputs into the \textbf{29-dimensional relative Cartesian action space}. This dataset consists of approximately \textbf{200 hours} of \algabbr demonstrations.
    
    \item \textbf{Task-Specific Finetuning.} After this adaptation, we further finetune the model end-to-end with \textit{task-specific datasets}. This stage ensures that the model not only operates correctly in the relative Cartesian space but also achieves \textbf{maximal performance} on the target tasks.
\end{enumerate}

During both training and inference, the selected \memabbr head keyframe images are integrated directly into the Pali-Gemma vision-language model as additional context image tokens without requiring changes to the core network or introducing learned memory modules.

\subsection{Policy Deployment}

The action control frequency is set to 25Hz while the policy model asynchronously inferences at approximately 40 ms latency on an RTX 5090 GPU. The policy outputs a horizon of 40 future end-effector and head poses and gripper commands. We adopt temporal chunk ensembling~\cite{zhao2023learning} as our default control strategy. 

To map these targets (left hand, right hand, and head) into the robot's specific joint configurations, we employ a differentiable inverse kinematics (IK) solver (Pyroki) \cite{kim2025pyroki}. Unlike analytical IK, which fails or returns nulls on unreachable poses, a differentiable approach treats IK as a weighted cost-minimization problem. By optimizing for end-effector pose error alongside posture regularization, the system achieves "graceful degradation" reaching as close as physically possible to the demonstrated pose rather than experiencing execution errors. This ensures robust transfer of unconstrained human demonstrations to robot embodiments with varying kinematic limits without requiring manual trajectory filtering or retraining. 

We maintain hardware modularity by using the Viser~\cite{yi2025viser} toolkit for high-level scene-graph management. Rather than authoring a monolithic URDF, we programmatically align the mounting interfaces of the Rainbow RBY1 and I2RT YAM actuated neck within a global coordinate frame. During deployment, two independent Pyroki processes resolve the torso/arm and head trajectories in parallel. This decoupled architecture ensures the 29D action representation remains platform-agnostic, facilitating zero-shot transfer to heterogeneous hardware configurations without retraining.

\section{Experiments}

\begin{figure*}[t]
  \centering
  \includegraphics[width=1.0\linewidth]{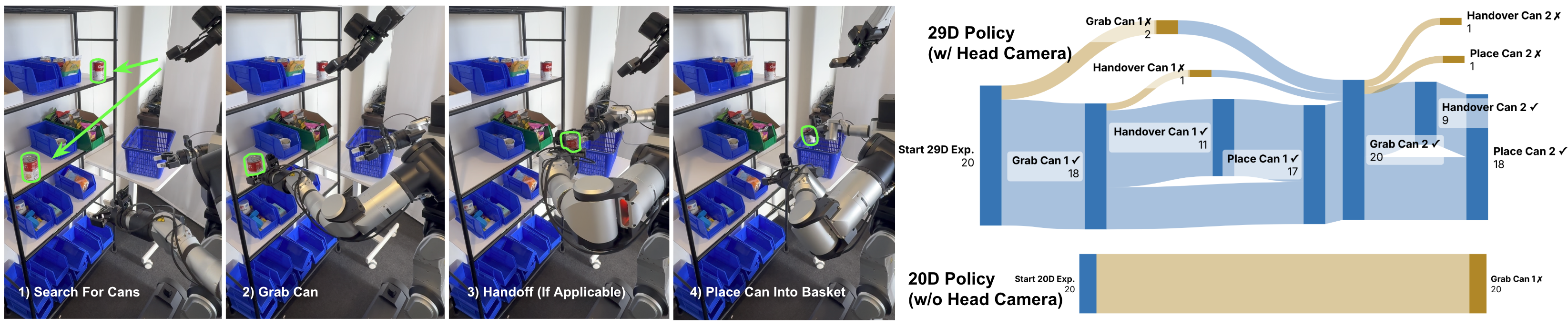}
  \caption{
  \textbf{Shelf Task Rollout Sequence}. \textit{(Left).} The images show a real 29D policy evaluation rollout where the robot (1) scans across multiple shelf tiers to locate target cans, (2) reaches and grasps the selected item, (3) performs a mid-air inter-gripper handoff, and (4) places the item into a shopping basket, then repeats on the remaining can. \textit{(Right).} The Sankey diagrams show failure modes for the 29D active-head, whole-body retargeted policy compared to the 20D wrist camera-only policy.
  }
  \label{fig:shelf_picking}

  \vspace{-10pt}
\end{figure*}

We evaluate \algabbr on a suite of real-world manipulation tasks designed to test the impact of active head retargeting and memory-augmented policies. 
Our evaluation focuses on two key aspects: (1) the role of explicit head pose retargeting and head-camera observations in enabling robust wide-spanning bimanual manipulation, and (2) the necessity of \memabbr for handling tasks requiring visual memory under partial observability. 
All experiments are conducted on the robot platform described in Sec \S\ref{sec:platform}, with policies trained from only demonstrations collected from our VR-based data collection device, with zero teleoperated on-embodiment data.

\subsection{Data Collection and Policy Training}
For each task in the sections below, 1-1.5 hours of in-domain task-specific data was collected on the \algabbr device described in Sec. \S\ref{sec:hardware}. Evaluation policies were trained for 40k steps, taking approximately 50 hours each.

\subsection{Searching Tasks}

\subsubsection{Task Setup}
The searching experiments evaluate the \algabbr framework in robot capability to localize and manipulate a target object in a wide workspace where the target may initially lie outside the field of view of both head and wrist cameras. 
Two searching tasks are considered: 

\textbf{(1) Tabletop Search:} 
A soup can is placed within a large randomization distribution on a 30” $\times$ 60” table along with up to eight distractor objects. Importantly, the soup can can may start outside of the immediate field of view of the initalized robot.
The robot must locate the soup can and place it into a designated organizing bin.  
When the target and bin reside on opposite sides of the table, operators are instructed to execute a transfer maneuver: pick with the closer hand, place the object temporarily at the center, then regrasp with the opposite hand to complete the placement.  
This produces natural and diverse bi-manual behaviors that emphasize the need for wide-spanning search and hand-to-hand coordination.

\textbf{(2) Shelf Search:} 
A tall shelving unit (height $\approx$ 2.4\,m) contains distractor items and 2 soup cans, which may be placed at any shelf location at two tiers near shoulder height.  
The robot must identify the target cans and place it in a shopping basket located on a nearby table to the right.
To succeed, the robot must search vertically and horizontally, and additionally execute a precise inter-gripper handoff when target objects are picked from the far side of the shelf.

\subsubsection{Results}
We compare and report two policy configurations:
\begin{itemize}
    \item \textbf{29D Policy:} Includes head SE(3) action outputs for active head actuation and head-camera images.
    \item \textbf{20D Policy:} Excludes head SE(3) action outputs and excludes head-camera images, using only wrist-camera observations and gripper SE(3) commands.
\end{itemize}

On the tabletop task, the 29D policy achieves a success rate of \textbf{36/40}, outperforming the 20D wrist camera-only policy, which achieves \textbf{29/40}.  
Failure analysis shows that the 20D policy struggles primarily in wide-spanning scenarios requiring hand transfers, often failing to coordinate across the workspace due to incomplete scene context within only the wrist-views.  
By contrast, the 29D policy leverages the operator's natural pre-attentive head motion: as operators look toward the placement location before moving their hands, head-centering ensures that the placement location is already within the observation context for the policy model. 

To further isolate the role of active head motion, we introduce an additional experiment in which the model is still provided with a head-camera image stream, but the ability to dynamically reposition the viewpoint is removed by fixing the head target.
Performance drops achieving only \textbf{2/20} successful trials. Without active head control, the policy struggles heavily with object grasping and placement accuracy.

On the shelf task, the benefits of active head modeling are even more pronounced.  
The 29D policy achieves \textbf{35/40} success points (each can placed correctly counts as one point), while the 20D policy achieves \textbf{0/40}.  
Without head pose and gaze retargeting, the 20D model fails immediately, unable to localize off-screen targets or coordinate vertical and lateral search motions.  
This demonstrates that natural operator head motion provides essential cues for planning long-range reaching and handoffs without requiring explicit operator instruction.

\subsubsection{Discussion}
These results confirm that head pose retargeting and active head image observations are critical for bridging the embodiment gap in wide-spanning manipulation.  
Without them, the robot cannot reason about objects outside the initial wrist camera view, leading to catastrophic failure on tasks involving search or cross-workspace coordination.

\subsection{Memory Tasks}

\begin{figure*}[t]
  \centering
  \includegraphics[width=1.0\linewidth]{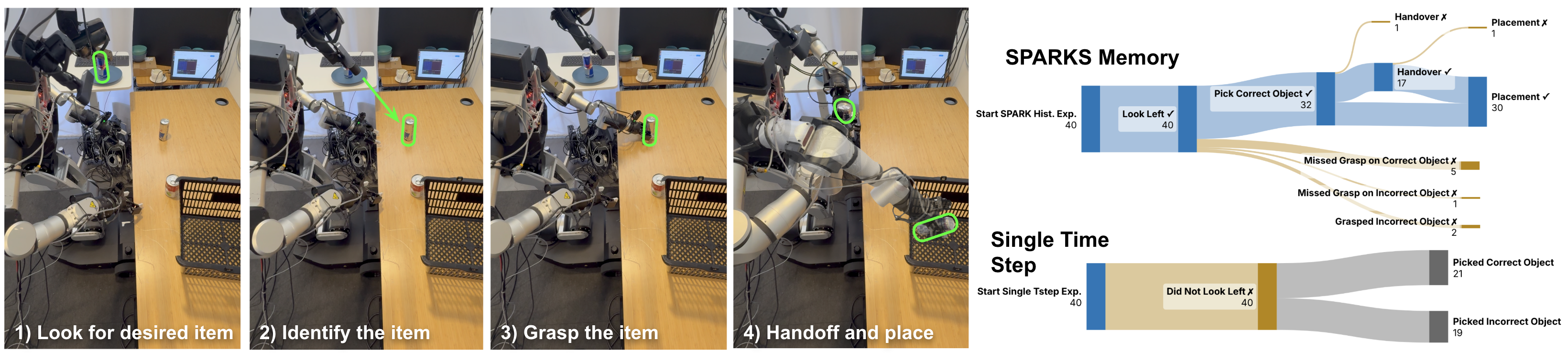}
  \caption{
  \textbf{Memory Task Rollout Sequence}. \textit{(Left).} The images show a real 29D policy evaluation rollout where the robot starts facing the table, then must (1) look left to identify the desired picking item, (2) look back to the table, (3) grasp the correct item (4) place the item into the bin with a mid-air handover if necessary. \textit{(Right).} The Sankey diagrams show failure modes for policies leveraging SPARKS compared to single time-step conditioning.
  }
  \label{fig:envs2}
  
  \vspace{-10pt}
\end{figure*}

\subsubsection{Setup}
The memory task evaluates whether \memabbr enables the robot to maintain spatial memory across spatial and temporal gaps.  
The setup consists of a forward table with two objects (a soup can and a drink can) and a placement basket on the right.  
A second side table, positioned at a 90$^\circ$ angle to the left and \emph{initially out of view}, contains either the soup can or the drink can (but never both).  

The task proceeds as follows: the robot must first \emph{look left} to inspect the side table, remember which object is there, then return to the forward table and pick and place the correct item into the basket. In the case that the correct object is on the left side, the operator is instructed to perform an inter-gripper handover before completing the placement into the shopping basket to the right

This design explicitly requires maintaining a temporally persistent memory of the visual information acquired during the initial head turn.

\subsubsection{Results}
We compare a baseline single-timestep policy (no memory) to our \memabbr memory-augmented policy:
\begin{itemize}
    \item \textbf{Single-Timestep Policy:} Conditions only on the current head and wrist camera images.
    \item \textbf{\memabbr-Augmented Policy:} Selects and retains past frames using the \memabbr keyframe selector, providing a compact history buffer as additional input.
\end{itemize}

The single-timestep policy achieves a success rate of \textbf{21/40}, which is near random chance.  
Qualitative analysis shows that this policy fails to look left, instead directly picking from the forward table based on an ambiguous current view.
By contrast, the \memabbr-augmented policy achieves \textbf{31/40} success, consistently looking left, adding the keyframe to memory, and then using it to disambiguate which object to select after the side table leaves the field of view.

\subsubsection{Discussion}
This experiment demonstrates the necessity of memory for tasks involving occluded or off-screen observations.  
Without \memabbr, the policy is forced into a Markovian assumption, treating the problem as if all relevant information is visible at once, which leads to high failure rates. 

Together, these findings validate \algabbr as a scalable approach for closing the embodiment gap and enabling complex whole-body manipulation policies trained entirely from egocentric human demonstrations. Notably, our method required neither visual augmentation nor on-embodiment data. We hypothesize that the perspective changes introduced by the actuated head—as opposed to a fixed overhead camera—encourage the policy to learn robust visual features that ignore irrelevant context and focus on task-relevant information. This underscores the importance of active perception in imitation learning and suggests that explicit head modeling can reduce dependence on data augmentation.

\section{Conclusion}

While EgoMI significantly narrows the embodiment gap, several limitations remain. The system is heavy and difficult to use for long durations for certain users. Physical embodiment mismatches also persist: for example, the robot’s actuated head can move beyond natural human ranges, so retargeting may constrain performance. \memabbr, though effective, uses a fixed scoring heuristic and does not adaptively update memory; more intelligent conditioning mechanisms could further improve performance.

EgoMI enables learning active vision and whole-body manipulation from egocentric human demonstrations. By synchronizing head and hand motion, leveraging spatial memory with \memabbr, EgoMI achieves embodiment transfer to real robots without additional robot data. These results highlight egocentric demonstration as a scalable path for bridging the human–robot embodiment gap and enabling more general robot behavior.










\bibliographystyle{IEEEtran}
\bibliography{references}  

\end{document}